\documentclass[runningheads]{llncs}
\authorrunning{M. M. Liu et al.}  
\titlerunning{OraPlan–SQL}
\usepackage{graphicx}
\usepackage{wrapfig}
\usepackage{booktabs}
\usepackage{graphicx}
\usepackage[inkscapelatex=true]{svg}  

\usepackage{graphicx}
\usepackage{amsmath}
\usepackage{CJKutf8}

\usepackage{float}  
\usepackage{threeparttable} 
\usepackage{tcolorbox}

\begin{document}

\title{OraPlan--SQL: A Planning-Centric Framework for Complex Bilingual NL2SQL Reasoning}

\renewcommand{\thefootnote}{\fnsymbol{footnote}}
\author{Marianne Menglin Liu$^*$ \and Sai Ashish Somayajula$^*$ \and Syed Fahad Allam Shah \and Sujith Ravi \and Dan Roth}
\institute{Oracle AI \\
\{marianne.liu, ashish.somayajula, syed.fahad.allam.shah, sujith.ravi, 
dan.roth\}@oracle.com}
\renewcommand{\thanks}[1]{\footnotetext{#1}}
\thanks{$^*$Equal contribution; authors listed in no particular order.}
\maketitle

\linespread{0.95}

\begin{abstract}

We present \textsc{OraPlan--SQL}, our system for the \textsc{Archer} NL2SQL Evaluation Challenge 2025, a bilingual benchmark requiring complex reasoning such as arithmetic, commonsense, and hypothetical inference. \textsc{OraPlan--SQL} ranked first, exceeding the second-best system by more than 6\% in execution accuracy (EX), with 55.0\% in English and 56.7\% in Chinese, while maintaining over 99\% SQL validity (VA). Our system follows an agentic framework with two components: Planner agent that generates stepwise natural language plans, and SQL agent that converts these plans into executable SQL. Since SQL agent reliably adheres to the plan, our refinements focus on the planner. Unlike prior methods that rely on multiple sub-agents for planning and suffer from orchestration overhead, we introduce a \emph{feedback-guided meta-prompting} strategy to refine a single planner. Failure cases from a held-out set are clustered with human input, and an LLM distills them into corrective guidelines that are integrated into the planner’s system prompt, improving generalization without added complexity. For the multilingual scenario, to address transliteration and entity mismatch issues, we incorporate entity-linking guidelines that generate alternative surface forms for entities and explicitly include them in the plan. Finally, we enhance reliability through plan diversification: multiple candidate plans are generated for each query, with the SQL agent producing a query for each plan, and final output selected via majority voting over their executions.

\end{abstract}

\section{Introduction}
Natural language to SQL (NL2SQL) tasks test the ability of language models to bridge natural language understanding, structured reasoning, and symbolic execution \cite{spider2018,seq2sql2017}. 
The Archer NL2SQL Evaluation Challenge~\cite{zheng2024archer} advances this direction by introducing a bilingual benchmark that emphasizes complex reasoning beyond surface-level mapping. Queries involve arithmetic computation, commonsense inference, and hypothetical scenarios, making the task significantly more challenging than standard semantic parsing benchmarks \cite{pan2023knowledge,liu2025logiccat}.
 Performance is evaluated along two dimensions: SQL validity (VA) and execution accuracy (EX), which together measure both syntactic correctness and semantic faithfulness: VA is the proportion of syntactically valid SQL queries, while EX is the proportion of generated queries whose execution results exactly match the gold answers~\cite{zheng2024archer}. Goal of this work is to advance Text-SQL generation by improving on this benchmark.

\begin{figure*}[t]
  \centering
  \includegraphics[width=\textwidth]{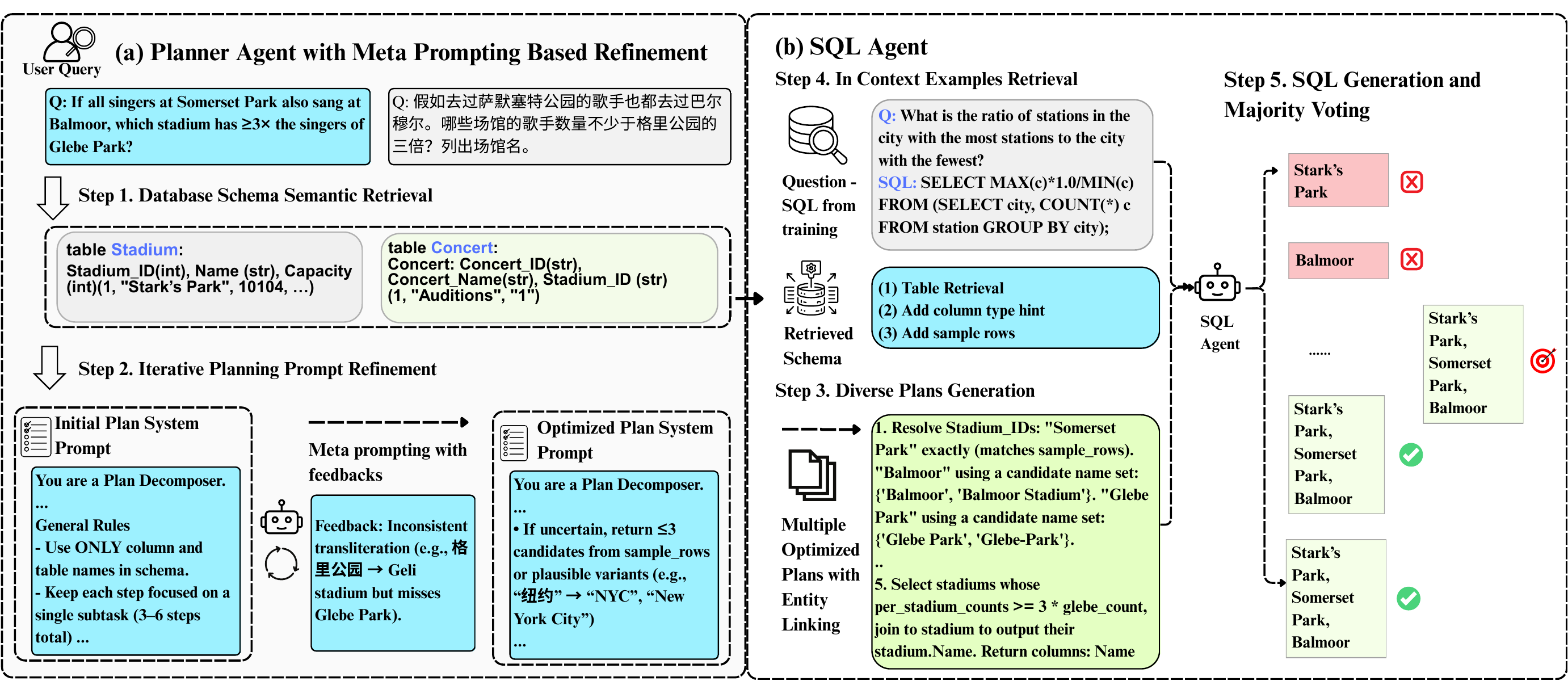}
  \caption{(a) A Planner Agent with an optimized system prompt, incorporating meta-prompting and entity linking to resolve transliteration and mismatch issues. Diverse plans are generated for robustness. (b) A SQL Agent processes these plans, using retrieved schema and in-context examples to enhance quality, with majority voting selecting the final output from the diverse SQL candidates.}
  \label{fig:oraplan-arch}
\end{figure*}

Prior work has advanced Text-to-SQL generation \cite{turn0academia32}, yet significant limitations remain in handling diverse reasoning scenarios \cite{turn0search1,turn0search2}. For example, the query ``If the number of students in the CS department increased by 10\% last year, how many are currently enrolled?'' requires combining hypothetical reasoning with database retrieval. Direct Input $\rightarrow$ Model $\rightarrow$ Output pipelines often misapply arithmetic or hallucinate joins, and such errors are difficult to diagnose or correct. ReFoRCE~\cite{deng2025reforce} improves robustness through relevant schema retrieval and self-refinement, but still treats Text-to-SQL as a direct mapping, leaving intent interpretation errors opaque. AgenticData~\cite{sun2025agenticdata} introduces a multi-agent planner that decomposes queries into sub-queries, each handled by specialized agents, enabling explicit reasoning but incurring orchestration overhead and brittle handoffs. We argue that both approaches overlook a middle ground: strengthening the system prompt of a single planner so that natural-language plans encode arithmetic, commonsense, and hypothetical reasoning. This approach mitigates the fragility of multi-agent pipelines while preserving explicit reasoning, offering a promising direction for more efficient and robust Text-to-SQL systems \cite{granado2025raise,pourreza2024chase}.

To address the limitations of prior works, we present \textsc{OraPlan--SQL}, a Text-to-SQL framework that integrates explicit planning with robust SQL generation. As shown in Figure~\ref{fig:oraplan-arch}, it consists of two agents, the planner agent that translates a user query into a stepwise plan in natural language and the SQL agent that generates executable SQL from the plan. Our key contributions are threefold: (1) We propose a feedback-guided meta-prompting approach for improving the system prompt of the planner: clustered failure cases from a held-out set, together with human feedback, are provided to an LLM, which then derives generalizable prompt guidelines that both correct observed errors and transfer to unseen cases. (2) We design multilingual entity-linking guidelines in the planner prompt to mitigate transliteration and mismatch issues~\cite{knight1997machine} by generating entity variants before SQL generation. (3) We enhance robustness through plan diversification by generating multiple candidate plans for each query, converting them into SQL, and selecting the final output via majority voting over their executions. These advances highlight the crucial role of planner prompt design in enhancing Text-to-SQL reasoning, especially in complex and multilingual settings. OraPlan--SQL ranked first on the Archer leaderboard, improving execution accuracy by more than $6\%$ over the second-best system (English: $55.0\%$, Chinese: $56.7\%$) while maintaining near-perfect validity ($>99\%$).

\section{Methodology}

In \textsc{OraPlan--SQL}, the Text-to-SQL mapping is decomposed into two stages: a \emph{planner agent} $f_{\text{plan}}$ that generates an intermediate natural language plan $P$, and a \emph{SQL agent} $f_{\text{sql}}$ that converts this plan into an executable SQL query $\mathcal{S}$:
\[
f_{\text{plan}}: (Q, \mathcal{D}) \mapsto P, \quad
f_{\text{sql}}: (P, \mathcal{D}) \mapsto \mathcal{S},
\]
where $Q$ is the input query and $\mathcal{D}$ is the database schema. The overall mapping is expressed as
\[
f = f_{\text{sql}} \circ f_{\text{plan}}.
\]
This decomposition introduces an explicit intermediate plan $P$ to make reasoning steps transparent, while the SQL agent ensures accurate SQL generation. In the following sections, we describe the optimizations applied to each component.

\subsection{Enriching Prompts with Schema Retrieval and In-Context Examples}
Following ReFoRCE~\cite{deng2025reforce}, we avoid appending the full database schema, which degrades performance. Instead, we use an embedding-based retriever to select the top-$k$ relevant schema elements, integrated into the planner and SQL agent's system prompts to ensure schema-aware generation and reduce hallucinations. We also apply in-context learning (ICL) by incorporating semantically similar training queries and gold SQL pairs into the SQL agent’s system prompt. While ICL is limited to the SQL agent due to the lack of ground-truth plans in the training set, extending it to the planner is a potential future direction. 

\subsection{Planner Agent and Meta-Prompting Based Refinement}
The planner agent is central to OraPlan–SQL, converting a natural language query and retrieved schema into a step-by-step plan. This plan is then passed to the SQL agent, which reliably produces executable SQL. Most errors stem from the planner, making its improvement crucial for system performance.

Instead of relying on multiple specialized sub-agents as in AgenticData~\cite{sun2025agenticdata}, we adopt a simpler (yet effective) approach based on analyzing planner outputs on a held-out set to identify common failure modes. These include errors in temporal reasoning, arithmetic, and counterfactual cases. With human input, the errors are clustered, and an LLM derives corrective guidelines that are incorporated into the planner’s system prompt to address these errors and generalize to related unseen cases. We refer to this process as \emph{feedback-guided meta-prompting}. It is \emph{meta-prompting} because the LLM does not generate task outputs (plans) directly, but instead produces higher-level guidelines that shape the planner’s own system prompt. It is \emph{feedback-guided} because it explicitly leverages both human input and systematic error analysis. By iteratively refining the planner’s system prompt with these derived guidelines, the agent improves accuracy of planning across diverse reasoning categories, offering a lightweight yet effective alternative to complex multi-agent planning.

In the bilingual setting, particularly for Chinese, we observed transliteration issues where proper nouns in the query did not align with database entries. To address this, we integrate entity-linking guidelines into the planner’s system prompt. These guidelines instruct the planner to generate alternative surface forms for all entities encountered in the query (e.g., ``NYC'', ``New York City''), which are then directly incorporated into SQL generation to improve entity matching and accuracy. Detailed guidelines derived from our meta-prompting approach are presented in Appendix~\ref{sec:error_cases}.

\subsection{SQL Agent}
The SQL agent converts each natural language plan into an executable query. Its prompt is augmented with the top-$5$ retrieved schema elements and semantically similar in-context examples, along with minimal guidelines tuning. Together, these enhancements improve schema awareness, ensure syntactic correctness, and keep the generated SQL faithful to the plan.

\subsection{Generate SQL for Diverse Plans and Majority Voting}
To enhance robustness, we generate diverse plans by setting the planner’s temperature to 0.7. Each resulting plan is passed to the SQL agent, executed, and the final answer is determined through majority voting on sql execution results, reducing the impact of errors from any single plan.

\section{Experiments}
\subsection{Main Results}
\label{sec:main-results}
OraPlan--SQL achieves \textbf{state-of-the-art bilingual performance} on \textsc{Archer} Challenge, substantially outperforming the 2025 runner-up HIT--SCIR. On \textbf{test set}, OraPlan--SQL improves execution accuracy (EX) by \textbf{6.3 points in English} (54.96\% vs.\ 48.66\%) and \textbf{12.6 points in Chinese} (56.67\% vs.\ 44.08\%). These results highlight the effectiveness of our framework, which leverages guidelines derived through meta-prompting to refine the planner and enhance execution accuracy. Furthermore, OraPlan--SQL achieves nearly identical accuracy across English and Chinese, \textbf{closing the cross-lingual gap} observed in prior systems.

\begin{table}[H]
\centering
\small
\begin{threeparttable}
\caption{Execution Accuracy (EX, \%) on the Archer NL2SQL Test set: OraPlan--SQL versus 2025 evaluation competitors. EN refers to English and ZH refers to Chinese.}
\label{tab:leaderboard}
\renewcommand{\arraystretch}{1.1}
\setlength{\tabcolsep}{4.5pt} 
\small
\begin{tabular}{l l cc}
\toprule
\textbf{Team} & \textbf{Model} & EN & ZH \\
\midrule
\textbf{OraPlan--SQL} & GPT-5 (1st) & \textbf{54.96} & \textbf{56.67} \\
\midrule
HIT--SCIR & GPT-4o & 48.66 & 44.08 \\
Nirvana & Claude4+GPT-4o & 45.61 & 41.41 \\
JD5star-2025 & DeepSeek-V3.1-Think & 43.89 & 39.12 \\
KAMFU & Qwen3-235b-a22b & 43.32 & 37.40 \\
\bottomrule
\end{tabular}
\end{threeparttable}
\end{table}

\subsection{Ablation Studies}
We conduct ablation studies to assess the impact of individual components in \textsc{OraPlan--SQL} by removing or modifying one module and evaluating execution accuracy (EX) on the English Dev set. Results are shown in Table~\ref{tab:oraplan-ablation-model}.
 
\begin{itemize}
    \item \textbf{Planner Agent.}  To assess the value of explicit planning, we remove the planner agent and directly feed the query, retrieved schema, and ICL examples to the SQL agent. Performance drops from 72.12\% to 64.42\% ($-7.70$ points), confirming that intermediate natural language plans are crucial for accurate reasoning.

    \item \textbf{Guidelines from Meta-Prompting.}  We evaluate the effect of refining the planner’s system prompt with feedback-guided meta-prompting. Without including the generated guidelines in the planner's system prompt, performance is only 44.23\%, whereas incorporating guidelines boosts accuracy to 72.12\% on English queries. On Chinese queries, accuracy improves from 49.04\% to 79.81\%. These large gains highlight that the quality of the planner’s prompt is as important as the planning step itself.

    \item \textbf{Impact of Choice of Model.} We evaluate the effect of replacing GPT--5 with other model variants for both planning and SQL generation. The general-purpose \texttt{GPT-4o} performs the weakest (52.88\% EN, 45.19\% ZH), while reasoning-oriented models achieve substantially stronger results: \texttt{GPT-o3} reaches 67.31\% in both languages, and \texttt{GPT-5} further improves to 72.12\% (EN) and 79.81\% (ZH). Importantly, even with \texttt{GPT-4o}, OraPlan--SQL outperforms the second-best team HIT--SCIR on the dev set (52.88\% vs.\ 41.34\% in EN; 45.19\% vs.\ 31.73\% in ZH), underscoring the robustness of our framework across models.

    \item \textbf{ICL Examples for the SQL Agent.} Adding semantically similar in-context examples to the SQL agent’s prompt improves accuracy from 70.19\% to 72.12\% (+1.93 points), indicating a modest but positive impact on performance.

    \item \textbf{Majority Voting.} Finally, we study the impact of generating diverse plans and applying majority voting over their corresponding SQL executions. With a single plan, accuracy is 71.15\%; with multiple plans and majority voting, accuracy rises slightly to 72.12\%. While the gain is marginal, this step improves reliability by mitigating errors from individual plans. 
    
\end{itemize}

\begin{table}[H]
\centering
\caption{Ablation and model impact analysis of OraPlan--SQL on the Archer NL2SQL Dev set. 
Panel (A) shows ablation results on English (EN) with GPT--5, reporting execution accuracy (EX, \%) and relative change ($\Delta$EX). 
Panel (B) shows the effect of different generation models on EX (\%) for English (EN) and Chinese (ZH).}
\label{tab:oraplan-ablation-model}
\renewcommand{\arraystretch}{1.2}
\setlength{\tabcolsep}{5pt}
\begin{minipage}{0.62\linewidth}
\centering
\textbf{(A) Ablation Study (EN)} \\
\begin{tabular}{l c c}
\toprule
\textbf{Variant} & \textbf{EX (\%)} & $\Delta$\textbf{EX} \\
\midrule
\textbf{OraPlan--SQL} & \textbf{72.12} & \textbf{0.00} \\
w/o guidelines      & 44.23 & -27.89 \\
w/o planner agent   & 64.42 & -7.70 \\
w/o ICL examples    & 70.19 & -1.93 \\
w/o majority voting & 71.15 & -0.97 \\
\bottomrule
\end{tabular}
\end{minipage}%
\hfill
\begin{minipage}{0.34\linewidth}
\centering
\textbf{(B) Model Impact} \\
\begin{tabular}{l c c}
\toprule
\textbf{Model} & \textbf{EN} & \textbf{ZH} \\
\midrule
gpt-4o & 52.88 & 45.19 \\
gpt-o3 & 67.31 & 67.31 \\
gpt-5  & \textbf{72.12} & \textbf{79.81} \\
\bottomrule
\end{tabular}
\end{minipage}
\end{table}

\section{Conclusion}
We introduced \textbf{OraPlan--SQL}, a planning-driven framework for Text-to-SQL that achieves state-of-the-art results on the Archer benchmark in both English and Chinese with near-perfect SQL validity.  

Our approach combines three innovations: (1) feedback-guided meta-prompting that clusters planner errors and distills corrective guidelines, (2) entity-linking guidelines to resolve transliteration and mismatch issues, and (3) plan diversification with consensus execution to improve robustness via majority voting. These contributions make OraPlan–SQL more effective in complex multilingual settings and provide a scalable, reliable solution for Text-to-SQL.

\newpage

\section{Appendix}

\subsection{Error Case Patterns and Planner Guidelines}
\label{sec:error_cases}
Through error analysis of planner outputs, we identified recurring failure modes and distilled targeted mitigation guidelines, illustrated with select cases.

\begin{itemize}
  \item \textbf{Arithmetic and Aggregation.}
  Planners often make mistakes in handling division, ranking, or projection scope.

  \emph{Guideline:}
  \begin{verbatim}
- When sums, averages, percentages, or totals are needed:
  • Break into steps (e.g., compute per-row value, 
    then aggregate with SUM/AVG/COUNT).
  • Percentages must be stated as formula: 
    `percentage = 100.0 * part / total`.
  • Guardrail: force float division (use 100.0 not 100).
  • If the question asks for a rate, ratio, or percentage, 
    the output may remain float/decimal.
  \end{verbatim}



  \item \textbf{Hypotheticals and Counterfactuals.}
  Planners sometimes conflate assumptions with actions, which results in ill-formed queries.

  \emph{Guideline:}
  \begin{verbatim}
- Trigger: if the question has "if… / assume… / 
  suppose… / hypothetically…".
- Always output the two lines first:
    • Counterfactual Condition: <the assumption as a filter>
    • Action After Condition: <task with any extra filters>

Decomposition rule:
1. Treat the assumption as a filter on the modified entity
   (entity-side filter).
    • Exclude that entity from the base set, then re-add
      under the assumption (avoid double counting).
2. Apply other filters from the rest of the sentence
   (population-side filters, e.g., occupation, rank, date).
3. Compute the requested metric under this modified dataset.
  \end{verbatim}

\end{itemize}

\subsection{Cross-Lingual Query Handling with Entity Linking: Translation vs. Direct Generation}

We evaluated two strategies for processing Chinese queries: (i) a \emph{translation-based pipeline} that translates queries into English before planning, and (ii) a \emph{direct generation pipeline} that takes Chinese queries as input and directly generates English plans. The direct generation approach achieved much higher execution accuracy (79.8\% vs. 55.8\%) because translation decouples the planner from the original linguistic context, where accurate entity resolution depends on native phrasing. Translation artifacts often introduced errors—for instance, \begin{CJK}{UTF8}{gbsn} 
格里公园 \end{CJK} (Glebe Park) became \emph{Geli Park}, and \begin{CJK}{UTF8}{gbsn} 约翰·尼斯尼克 \end{CJK} (John Nizinik) was rendered \emph{John Nisinik}—which propagated into planning and execution, causing the performance gap.

\bibliographystyle{splncs04}   
\bibliography{references}      

\begin{thebibliography}{10}
\providecommand{\url}[1]{\texttt{#1}}
\providecommand{\urlprefix}{URL }
\providecommand{\doi}[1]{https://doi.org/#1}

\bibitem{deng2025reforce}
Deng, M., Ramachandran, A., Xu, C., Hu, L., Yao, Z., Datta, A., Zhang, H.: Reforce: A text-to-sql agent with self-refinement, format restriction, and column exploration. In: ICLR 2025 Workshop: VerifAI: AI Verification in the Wild (2025)

\bibitem{turn0academia32}
Deng, N., Chen, Y., Zhang, Y.: Recent advances in text-to-sql: A survey of what we have and what we expect. arXiv preprint arXiv:2208.10099  (2022)

\bibitem{pan2023knowledge}
Dou, L., Gao, Y., Liu, X., Pan, M., Wang, D., Che, W., Zhan, D., Kan, M.Y., Lou, J.G.: Towards knowledge-intensive text-to-sql semantic parsing with formulaic knowledge. Proceedings of the 2022 Conference on Empirical Methods in Natural Language Processing pp. 5240--5253 (2022), \url{https://aclanthology.org/2022.emnlp-main.413.pdf}

\bibitem{granado2025raise}
Granado, F., Lotufo, R., Pereira, J.: Raise: Reasoning agent for interactive sql exploration. arXiv preprint arXiv:2506.01273  (2025), \url{https://arxiv.org/abs/2506.01273}

\bibitem{turn0search1}
Kanburoğlu, A.B.: Text-to-sql: A methodical review of challenges and models. Turkish Journal of Electrical Engineering and Computer Sciences  \textbf{32}(3), ~4077 (2024), \url{https://journals.tubitak.gov.tr/elektrik/vol32/iss3/4/}

\bibitem{knight1997machine}
Knight, K., Graehl, J.: Machine transliteration. arXiv preprint cmp-lg/9704003  (1997)

\bibitem{liu2025logiccat}
Liu, T., Mao, X., Zan, H., Zhang, D., Li, Y., Liu, H., Kong, L., Hou, J., Li, R., Li, Y., Zheng, A., Zhang, Z., Zhewei, L., Zhang, K., Peng, M.: Logiccat: A chain-of-thought text-to-sql benchmark for complex reasoning. arXiv preprint arXiv:2505.18744  (2025), \url{https://arxiv.org/abs/2505.18744}

\bibitem{pourreza2024chase}
Pourreza, M., Li, H., Sun, R., Chung, Y., Talaei, S., Kakkar, G.T., Gan, Y., Saberi, A., Ozcan, F., Arik, S.O.: Chase-sql: Multi-path reasoning and preference optimized candidate selection in text-to-sql. arXiv preprint arXiv:2410.01943  (2024), \url{https://arxiv.org/abs/2410.01943}

\bibitem{turn0search2}
Seek.ai: Text2sql: An overview of past, present and future (2024), \url{https://www.seek.ai/blog/a-survey-of-text-to-sqls-past-present-and-future}

\bibitem{sun2025agenticdata}
Sun, J., Li, G., Zhou, P., Ma, Y., Xu, J., Li, Y.: Agenticdata: An agentic data analytics system for heterogeneous data. arXiv preprint arXiv:2508.05002  (2025)

\bibitem{spider2018}
Yu, A., Yin, Y., Choi, E., et~al.: Spider: A large-scale, high-quality benchmark for complex and cross-domain semantic parsing and text-to-sql tasks. In: Proceedings of the 2018 Conference on Empirical Methods in Natural Language Processing (EMNLP). pp. 3581--3591. Association for Computational Linguistics (2018), \url{https://www.aclweb.org/anthology/D18-1371/}

\bibitem{zheng2024archer}
Zheng, D., Lapata, M., Pan, J.Z.: Archer: A human-labeled text-to-sql dataset with arithmetic, commonsense and hypothetical reasoning. arXiv preprint arXiv:2402.12554  (2024)

\bibitem{seq2sql2017}
Zhong, V., Xiong, C., Socher, R.: Seq2sql: Generating structured queries from natural language. In: Proceedings of the 2017 Conference on Empirical Methods in Natural Language Processing (EMNLP). pp. 423--433. Association for Computational Linguistics (2017), \url{https://www.aclweb.org/anthology/D17-1042/}

\end{thebibliography}

\end{document}